\documentclass[10pt,twocolumn,letterpaper]{article}

\usepackage{wacv}
\usepackage{times}
\usepackage{epsfig}
\usepackage{graphicx}
\usepackage{amsmath}
\usepackage{amssymb}
\usepackage{multirow}
\usepackage[accsupp]{axessibility}  

\usepackage{biblatex}
\graphicspath{ {./images/} }
\def\wacvPaperID{967} 

\wacvfinalcopy 

\ifwacvfinal
\fi

\ifwacvfinal
\usepackage[breaklinks=true,bookmarks=false]{hyperref}
\else
\usepackage[pagebackref=true,breaklinks=true,colorlinks,bookmarks=false]{hyperref}
\fi

\begin{document}

\title{Generalized Facial Manipulation Detection with Edge Region Feature Extraction}

\author{Dong-Keon Kim\\
Sungkyunkwan University\\

{\tt\small kdk1996@skku.edu}
\and
Kwangsu Kim\\
Sungkyunkwan University\\

{\tt\small kim.kwangsu@skku.edu}
}

\maketitle

\ifwacvfinal
\thispagestyle{empty}
\fi

\begin{abstract}
   This paper presents a generalized and robust face manipulation detection method based on the edge region features appearing in images. Most contemporary face synthesis processes include color awkwardness reduction but damage the natural fingerprint in the edge region. In addition, these color correction processes do not proceed in the non-face background region. We also observe that the synthesis process does not consider the natural properties of the image appearing in the time domain. Considering these observations, we propose a facial forensic framework that utilizes pixel-level color features appearing in the edge region of the whole image. Furthermore, our framework includes a 3D-CNN classification model that interprets the extracted color features spatially and temporally. Unlike other existing studies, we conduct authenticity determination by considering all features extracted from multiple frames within one video. Through extensive experiments, including real-world scenarios to evaluate generalized detection ability, we show that our framework outperforms state-of-the-art facial manipulation detection technologies in terms of accuracy and robustness.
\end{abstract}
\section{Introduction}

As various AI-based generative models are developed \cite{kingma2014autoencoding,pmlr-v32-rezende14,NIPS2014_5ca3e9b1}, facial manipulated videos, so-called \textit{DeepFake}, are becoming a major social issue. These forged videos are becoming so sophisticated that it is difficult to determine the authenticity with the human eye. Moreover, the advent of these realistically synthesized videos allows malicious actions to threaten the creditability of public media and individual privacy \cite{vaccari2020deepfakes}, \eg{}, politicians' fake news, celebrity pornography. Therefore, effective detection of facial manipulation is a significant issue in the computer vision field.

\begin{figure}[t]
\begin{center}
\includegraphics[width=0.85\linewidth]{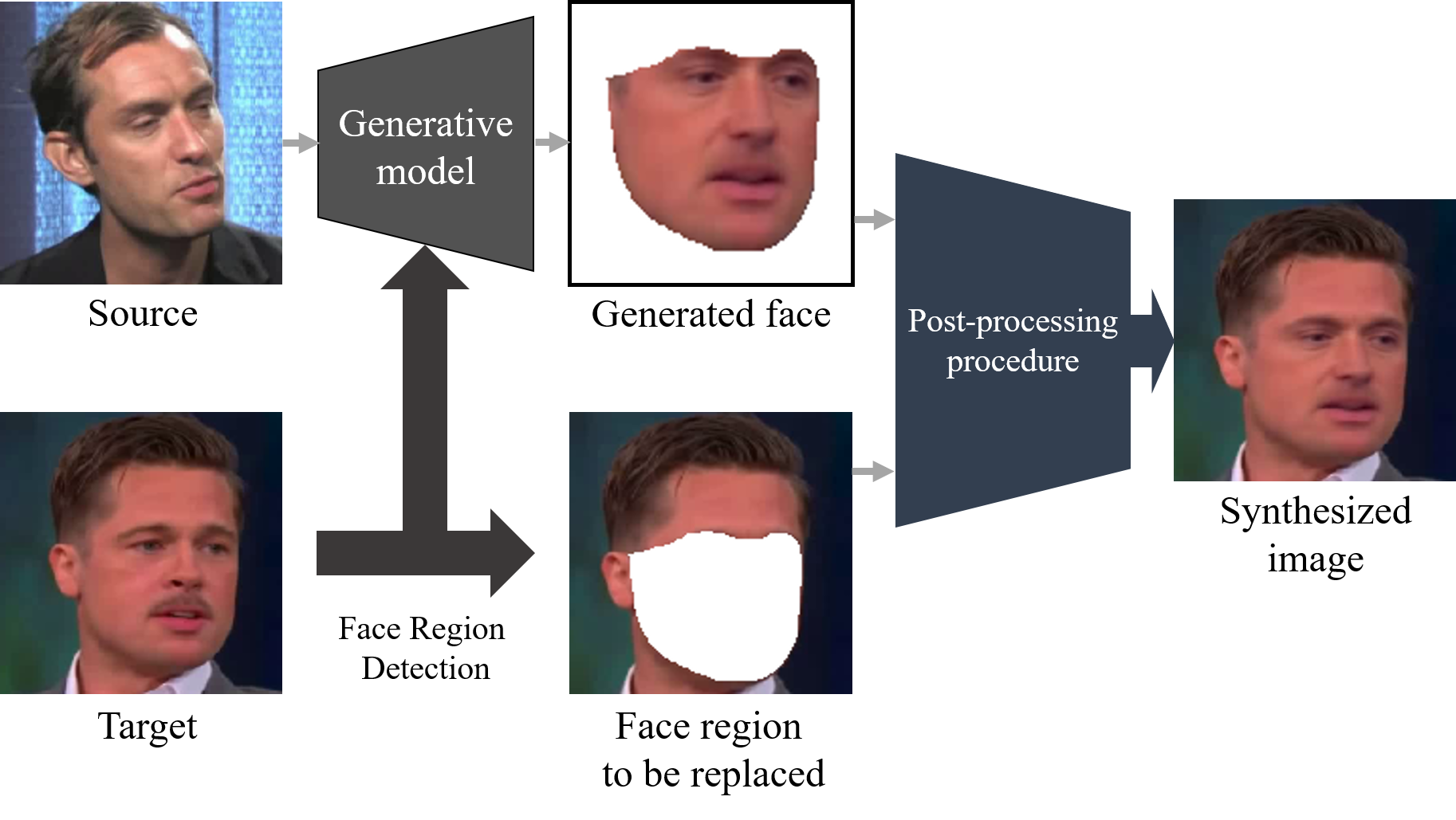}
\end{center}
   \caption{Summarized steps of general face manipulation procedure. An artificial image is created through a generative model with a source face image. Then, post processing is involved after image creation for a more realistic synthesis.}
\label{fig:fig1}
\label{fig:onecol}
\end{figure}

Most contemporary face synthesis algorithms generate a face shape (expression, mouth shape, etc.) learned by source images and pasting the shape into the face area of target images \cite{westerlund2019emergence}. Early \emph{DeepFake} videos in public datasets such as UADFV \cite{li2018ictu} and DeepfakeTIMIT \cite{korshunov2018deepfakes} commonly synthesized faces in target images without pixel-level correction, resulting in color awkwardness in the face edge region. In response, more advanced manipulation methods add post-processing procedures such as Gaussian blurring or Bi-linear interpolation to express the face-synthesized image more naturally \cite{li2020celebdf, mirsky2021creation}, as shown in Figure \ref{fig:fig1}. 

We observe that several modern face synthesis methods have the following common features: 1) Post-processing is done only on the facial part. Since the color correction is not performed on unrelated background parts, the inherent features of the background part are not damaged. 2) The face synthesis process does not involve the time concept. The intrinsic features vary in each frame, even with a slight change of pixels. However, the artificially synthesized videos do not highly consider the sequential changes. 

Based on these observations, we propose a generalized detection model of synthesized images. We mainly focus on a robust facial image forensic method through color distribution changes in the face-synthesizing process. Our proposed method pays attention to pixel-level features from the edge region in the entire, not only on the features appearing in the face, like other existing studies \cite{afchar2018mesonet, jeon2020fdftnet, 8682602}.

Also, a classification model including the temporal concept is introduced to improve detection performance. The latest studies \cite{qian2020thinking, yavuzkilic2021spotting} have already demonstrated that the framework that includes the time-domain properties has a much better forgery detection performance. Existing studies use known networks such as SlowFast \cite{feichtenhofer2019slowfast} and Xception \cite{chollet2017xception} as backbone. However, these models are not suitable for handling very small-sized chronospatial inputs. Therefore, a 3D classification model based on DenseNet \cite{huang2017densely} is introduced for dealing with the extracted small-size edge region features spatiotemporally.

This paper has great potential to contribute to the field of face manipulation detection. More specifically, our framework is unique in that it reflects the edge attributes appearing throughout the whole image and interprets these features spatiotemporally. Several cross-experimental results show that our method outperforms other existing detection frameworks in accuracy and generalization ability.


\section{Previous Work}

In response to the rising threats of AI-generated facial forgeries, many researches \cite{du2019towards, marra2018detection, quan2018distinguishing, ding2020swapped, guera2018deepfake, li2018ictu} have been held to detect forged images. Existing studies devise their methods based on the abnormality revealed in the synthesized facial area. For example, Matern et al. \cite{matern2019exploiting} present a detection method based on visual artifacts in eyes and noses. Yang et al. \cite{yang2019exposing} suggest a \textit{DeepFake} detection methodology focusing on the inconsistency of head poses. While these works pay attention to the awkwardness features of synthesized faces, many recent works \cite{afchar2018mesonet,sabir2019recurrent,zhou2017two,zhou2018learning} presents detection methods with state-of-the-art artificial neural networks such as the recurrent neural network \cite{guera2018deepfake} and capsule network \cite{8682602}. All of these studies commonly focus on the forged image itself rather than the synthesis process. This approach inevitably exposes limitations in practical situations where it is necessary to determine the authenticity of video data made from various sources.

The facial manipulation detection should remain robust even with unseen forged data to apply the detection framework in real-world situations. The latest studies \cite{cozzolino2018forensictransfer,du2019towards, xu2021visual, dang2020detection, jiang2021practical} focus on the innate artifacts that appear in the face image synthesis process for a more practical detection framework. For instance, Face X-ray \cite{9157215} focuses on common blending procedures and shows their generalization ability with unseen datasets. SPSL \cite{liu2021spatial} analyzes forged images as a phase spectrum that emphasizes the abnormal features from the up-sampling procedure in generative models. These novel studies also show good performance in cross-dataset experiments, but they also have a problem of observing only the features of the face part. This limitation critically loses generality when cross-validating the source data with completely disparate datasets (\eg{} FaceForensics++ \cite{Rossler_2019_ICCV} for train, Celeb-DF \cite{li2020celebdf} for test). Our study further derives a better performance with more generality by comparing the extracted features between facial edges and background edges.

\section{Edge Region Feature}\label{fe}

Early research in image forensics \cite{farid2009survey} mentions that unforged natural images have unique fingerprints and artifacts. These traces are affected by external factors such as the surface of a camera lens, the direction of the light entering the camera, and the image processing procedure. Moreover, recent studies \cite{li2020identification, mccloskey2019detecting, yu2019attributing, marra2019gans} show that AI-based generative models do not address color component distortion or unnatural (artificial) imprints, resulting in imperfect reproduction of original imprints. Thus, the face-replacing and the color-correction process damage the raw image fingerprints across the facial boundary. We pay attention to this point by extracting the color difference features in both the facial part and the background area at the edge region.

\begin{figure}[t]
\begin{center}
\includegraphics[width=0.8\linewidth]{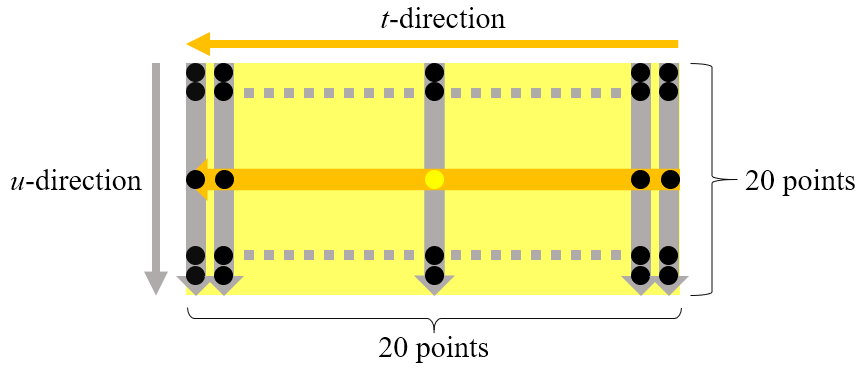}
\end{center}
   \caption{A brief illustration of edge window. Each point is spaced equally above the line segments. A center point in the window implies mid-point \(M_{i, i+1}\) of 2 neighboring landmark points, \(L_{i}\) and \(L_{i+1}\).}
\label{fig:fig3}
\end{figure}

\subsection{Facial Edge Region Feature}\label{fb}

\begin{figure*}
\begin{center}
\includegraphics[width=0.8\linewidth]{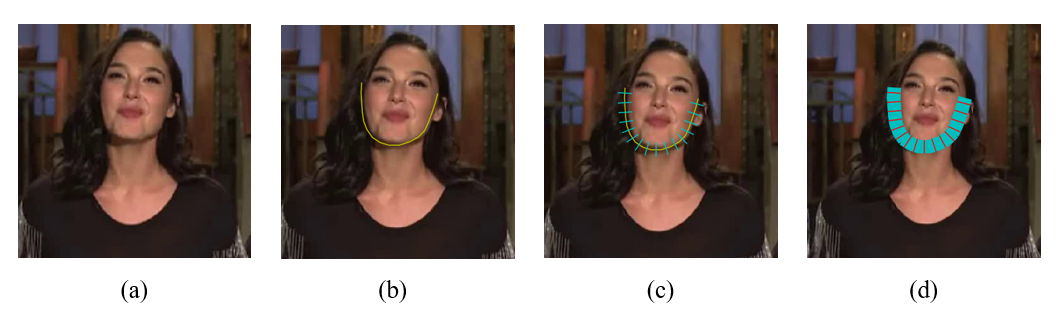}
\end{center}
   \caption{Facial boundary window generation pipeline. (a) Capture image \(I\) from video. (b) Get a facial boundary line with face landmark points detected from (a). (c) Make perpendicular line segment \(P_{i, i+1}\) with neighboring landmark points. (d) Create a set of horizontal lines \(V_{i, i+1}\) perpendicular to line segment \(P_{i, i+1}\) at equal intervals.}
\label{fig:fig4}
\end{figure*}

With the frame image \(I\) extracted from the video, the face area from \(I\) is detected with the frontal face detector in dilb \cite{king2009dlib}. Then the 68 feature points are extracted from the detected face with 68 face landmark shape predictors in dlib. Among them, we define the first 17 feature points representing the facial contours as facial boundary points, FBPs. We would like to create a line segment perpendicular to the facial boundary line to obtain color features along the face edge region. The \(x\)-coordinate and \(y\)-coordinate corresponding to the line segment element \(P_{i,i+1}\) are generated with the coordinates of two neighboring landmark points \(L_i\) and \(L_{i+1}\) in FBPs. The creation method follows parametric equations \eqref{eq1} and \eqref{eq2}.

\begin{equation}\label{eq1}
P_x (t)_{i,i+1} = \frac{(L_{i, y} - L_{i+1, y})\sqrt{S_I}}{D_{i,i+1}} t + M_{i,i+1,x}
\end{equation}

\begin{equation}\label{eq2}
P_y (t)_{i,i+1} = \frac{-(L_{i, x} - L_{i+1, x})\sqrt{S_I}}{D_{i,i+1}} t + M_{i,i+1,y}
\end{equation}

\noindent Here, \(t\) is a 20 integer parameter with a range of -10 to 9, from the inside to the outside of the face. \(D_{i,i+1}\) and \(M_{i,i+1}\) indicates distance and midpoint between two landmark points \(L_i\) and \(L_{i+1}\) each. Midpoint \(M_{i,i+1}\) is located on \(t\) = 0. Note that \(x\), \(y\) in subscript are \(x\)-coordinate and \(y\)-coordinate, respectively. \(S_I\) is a scale-factor for adjusting the length of the perpendicular line segment according to the size of the face in the image \(I\). \(S_I\) is obtained through the following equation \eqref{eq3}.

\begin{equation}\label{eq3}
S_I = \frac{\alpha}{A_I} 
\end{equation}

\noindent \(A_I\) is the convex area value covering the face contour created from the FBPs, which are obtained from image \(I\). \(\alpha\) is a tunable hyper-parameter that is related to the interval between line segment coordinates. By tuning with the scale-factor, the perpendicular line segment \(P_{i,i+1}\) passes through a specific range of the facial boundary, regardless of face size in the image.

With gathered line segments, we create window-shaped features to reflect the boundary region features spatially. A set of horizontal line segments \(V_{i, i+1}\), which are parallel to the facial boundary line, are considered. This horizontal line goes through the \(t\) point of obtained line segment \(P_{i,i+1}\). The concrete coordinates of horizontal line segments \(V_{i, i+1}\) are generated by following parametric equations \eqref{eq4} and \eqref{eq5}. 

\begin{figure}[t]
\begin{center}
\includegraphics[width=0.9\linewidth]{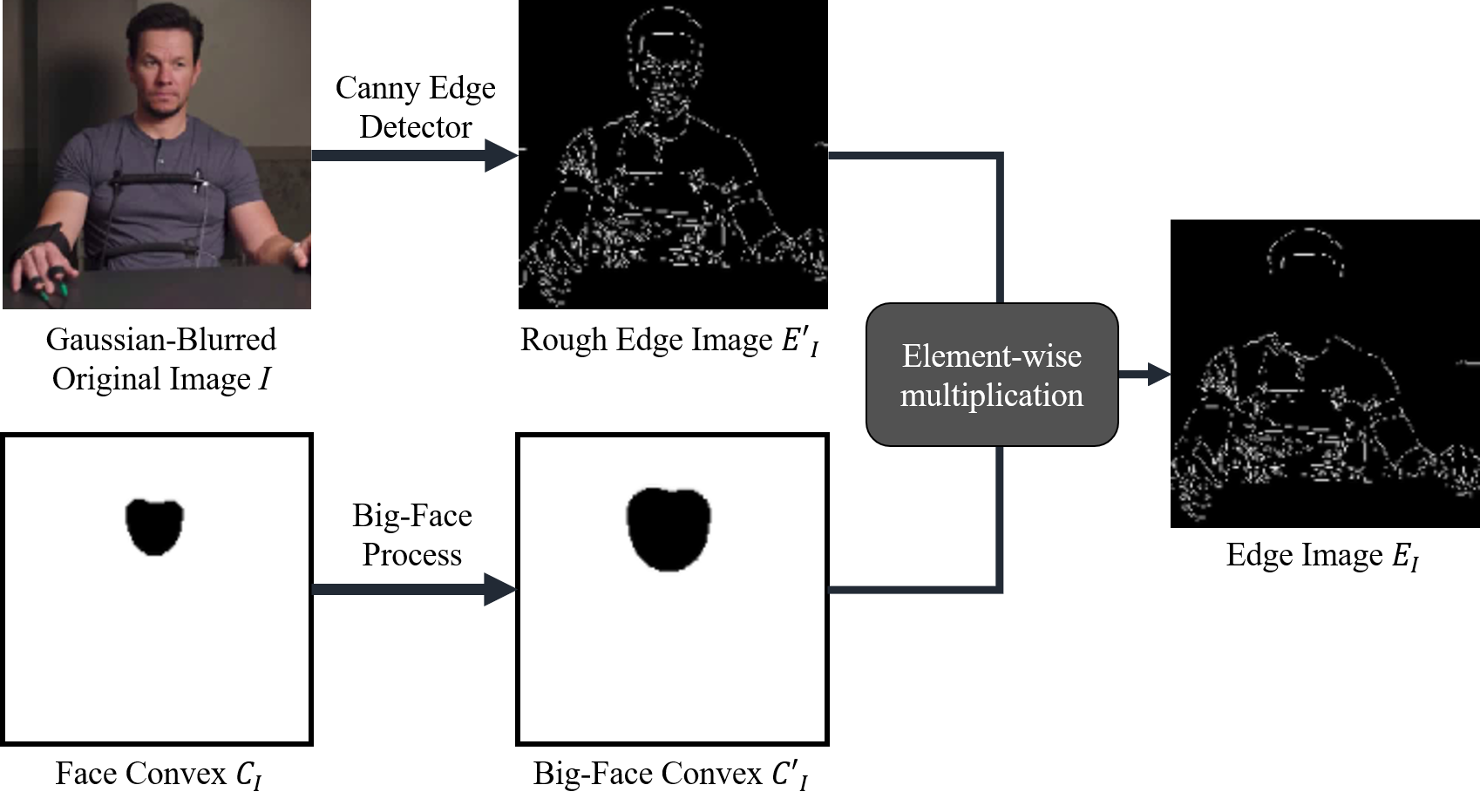}
\end{center}
   \caption{Background edge image generation procedure.}
\label{fig:fig5}
\end{figure}

\begin{equation}\label{eq4}
V_x(u,t)_{i,i+1} = \frac{(L_{i+1,x} - L_{i,x})\sqrt{2S_{I}}}{\beta}u +P_x(t)_{i,i+1}
\end{equation}

\begin{equation}\label{eq5}
V_y(u,t)_{i,i+1} = \frac{(L_{i+1,y} - L_{i,y})\sqrt{2S_{I}}}{\beta}u +P_y(t)_{i,i+1}
\end{equation}

Horizontal line segments are also represented by a parametric equation. Note that parameter \(u\) which has 20 integer parameters with range -10 to 9. \(\beta\) is a tunable hyper-paramter related to interval between horizontal line segment coordinates. The window-shaped 20\(\times\)20 corresponding coordinates are specified by a set of 20 horizontal lines. Figure \ref{fig:fig3} illustrates an example of an extracted edge window. We will refer to these 400 coordinate features as edge window \(W_{i}\), a set of horizontal lines \(V_{i, i+1}\) with two parameters \(t\) and \(u\). A total of 16 windows are created per single image \(I\) with facial boundary points (i.e., \(i \in \mathbb{N} \mid  1\leq i \leq 16\)). A summarized pipeline of window generation from a facial image is shown in Figure \ref{fig:fig4}. 

After creating the edge window \(W_{i}\), we compute the absolute difference of the RGB pixel values according to the parameter \(t\) to find out how the RGB pixel values in edge window \(W_{i}\) change from the inside to the outside of the face. These extracted 16 windows shaped 20\(\times\)19\(\times\)3 color value differences are \(F_{I}\), which is the facial boundary features of the image \(I\). Note that 20 refers to the number of \(u\)-direction coordinates in edge windows, 19 implies the modified parameter \(t\) (\(t=0\) is center) on which differentiated color value lie, and 3 indicates RGB.

\begin{figure*}
\begin{center}
\includegraphics[width=0.8\linewidth]{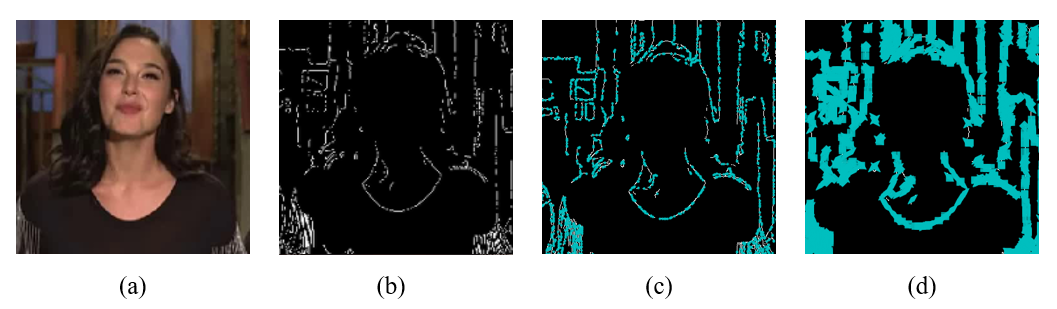}
\end{center}
   \caption{Background boundary feature extraction pipeline. (a) Capture image \(I\) from video. (b) Make a background edge image without facial boundary \(E_{I}\). (c) Sample 10\% points from total edges in the edge image \(E_{I}\). (d) Make vertical lines and horizontal lines in the same way as the facial line segments were extracted. }
\label{fig:fig6}
\end{figure*}

\subsection{Background Edge Region Feature}\label{bb}

Before extracting background edge points' features, we make a background edge image \(E_{I}\) from frame image \(I\). We firstly blur image \(I\) with 5\(\times\)5 Gaussian filter to removing noise of background and extract distinctive boundary parts. Then, the rough edge image \(E^{'}_I\) including edge points on the face is generated with the Canny edge detector \cite{canny1986computational}.

The facial edge points in \(E^{'}_I\) must be excluded to extract the natural fingerprint of the background edges. We considered convex image \(C_{I}\) that covers face contour made with the FBPs. \(C_{I}\) is a binary image with the same size as the original image \(I\) with inner points of convex are 0, and the outer points of convex are 1. A big-face convex image \(C^{'}_{I}\) is created by blurring \(C_{I}\) with 15\(\times\)15 Gaussian Filter and rounding down each value in blurred \(C_{I}\) to 0, except the pixels with the values of 1. We can get an edge image \(E_{I}\) that excludes facial edge points by multiplying \(C^{'}_{I}\) and \(E^{'}_I\) element-wise. A brief pipeline of making background edge image \(E_{I}\) is shown in Figure \ref{fig:fig5}.

With edge image \(E_{I}\), 10\% of the total edge points in \(E_{I}\) are randomly sampled to get statistical information that appears along background edges. Then background windows are extracted with sampled edge points in the same way as facial boundary features are, with \(L_{i}\) and \(L_{i+1}\) corresponding to the two nearest points of a sampled point. The extracted features are \(N\) windows shaped 20\(\times\)19\(\times\)3 where \(N\) is the number of sampled edge points in background edge image \(E_{F}\) while 20, 19, and 3 are equivalent to that mentioned in Section \ref{fb}. A summarized procedure of the background edge windows generation is shown in Figure \ref{fig:fig6}.

The direction of parameter \(t\) in the background edge windows from the \(E_{I}\) cannot be specified except that their boundary lines are in a vertical direction of edge points. For this reason, we fold the background edge feature and symmetrically unfold with the basis on \(t=0\) to remove the directional information in the edge window (\eg, all feature values in \(t=-9\) and \(t=9\) are averaged by the axis). Finally, the values of the N windows are averaged to get statistical information of the sampled N window features. A \(B_I\), which is the final background boundary feature of image \(I\), is the averaged background edge region features shaped 20\(\times\)19\(\times\)3.


\begin{figure*}
\begin{center}
\includegraphics[width=0.8\linewidth]{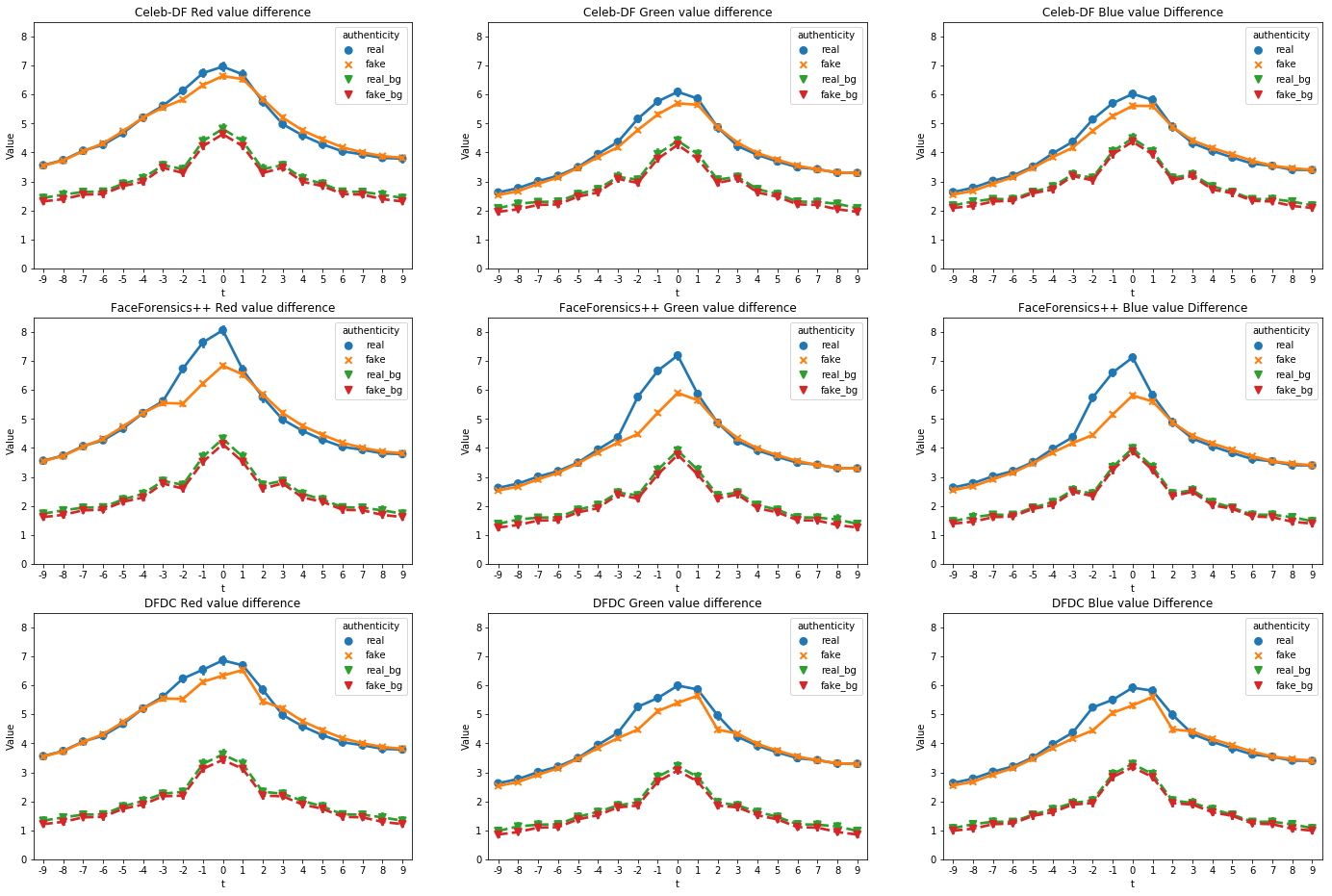}
\end{center}
   \caption{Color difference distribution of extracted features from 3 different datasets. (Celeb-DF, FaceForensics++ and DFDC)}
\label{fig:fig7}
\end{figure*}


\section{Analysis with Extracted Features}\label{aef}

We apply the proposed feature extraction method to face forensic datasets. Several datasets are adopted: FaceForensics++ (FF++) \cite{Rossler_2019_ICCV}, Celeb-DF \cite{li2020celebdf}, and DFDC \cite{dolhansky2019deepfake}. All of these datasets contain genuine video data and face manipulated video data forged from genuine ones. We randomly choose 750 genuine videos and 750 manipulated videos for FaceForensics++, and DFDC each. Exceptionally, the Celeb-DF dataset has less than 750 genuine video data, so we choose 563 genuine video data and 750 manipulated video data in the Celeb-DF dataset. Since the length of videos included in datasets varies, 24 consecutive frames at random timepoint are extracted.

The features extracted in Sections \ref{fb} and \ref{bb} are additionally post-processed for statistical analysis. The \(F_I\), which has 16 edge windows, is replaced with the averaged edge window. Then the edge region feature values of both face and background are averaged along the \(u\)-axis. The averaged \(F_I\) and \(B_I\), which both are shaped 19\(\times\)3, are the statistical data of the extracted edge region features. Here, 19 indicates the \(t\)-axis, which is the vertical direction of the boundary line, and 3 refers to RGB.

We plotted a graph to statistically analyze the change in the color difference along the vertical line of the edge regions. Figure \ref{fig:fig7} shows the dataset-wise distributions of facial features and background features along parameter \(t\)-direction. Note that boundary edge points are at \(t=0\). We observe a distribution difference between the facial features derived from the face manipulated image and those of the real image. There is a significant difference in distribution points close to \(t=0\), indicating the point along the facial boundary line.

In all three datasets, only the color distribution of the facial edge region has a distinct difference between the real and the fake. In addition, the difference between the color distribution of the forged facial boundary area and those of the background boundary area is similar regardless of the dataset. This observation shows that reflecting the facial information and background information together could help solve the generalization problem.

\section{Classification Model}

\begin{figure*}
\begin{center}
\includegraphics[width=0.85\linewidth]{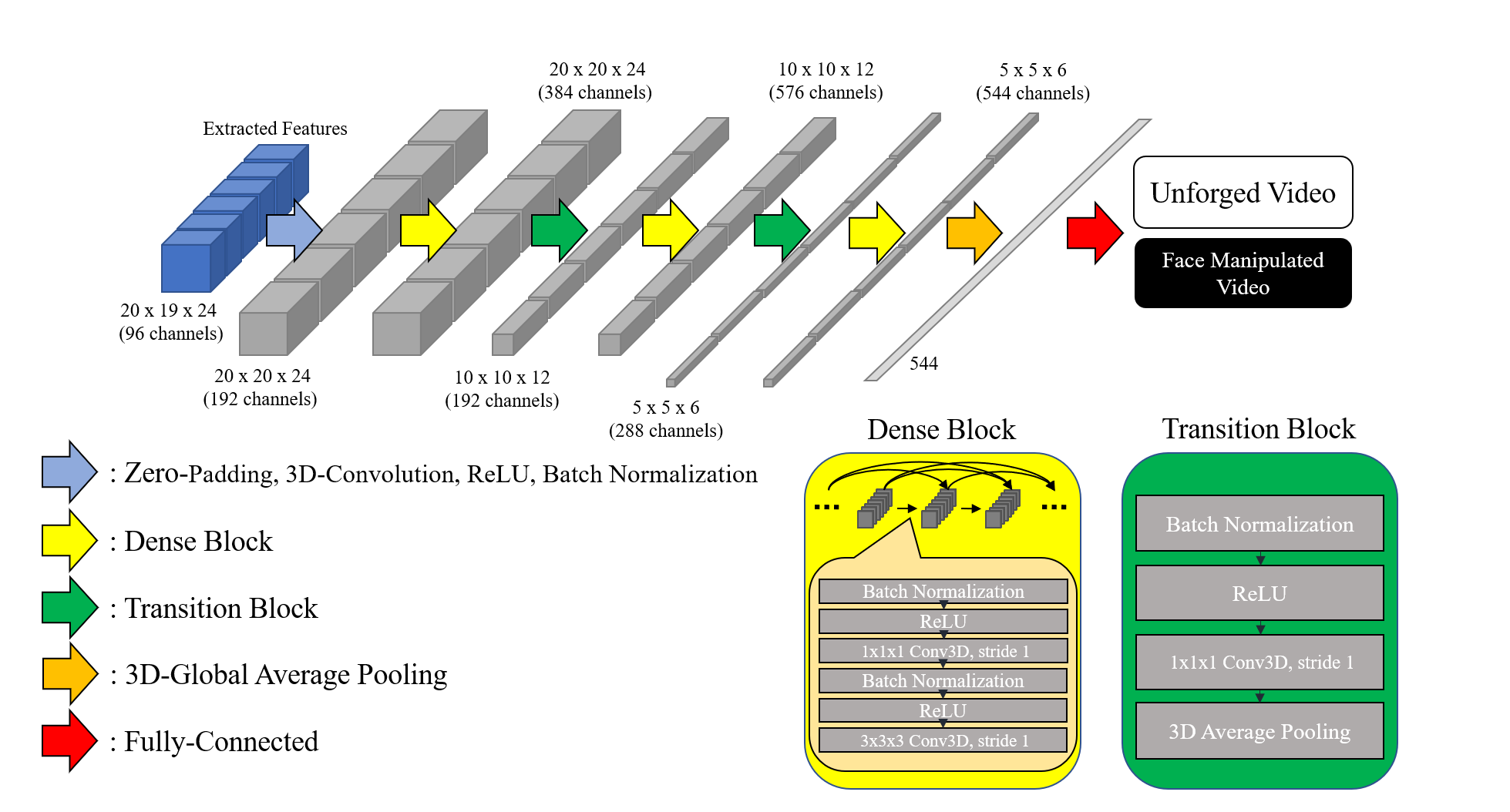}
\end{center}
   \caption{Illustrated structure of our proposed classification network, 3D-LightDenseNet. 3D-LightDenseNet extracts further distinctive spatio-temporal characteristics with frame-wise extracted edge region features and determines authenticity of input video.}
\label{fig:fig8}
\end{figure*}

Recent studies \cite{dang2020detection, 9157215, li2018exposing} attempted to find a more generalized image forensic solution by localizing the face synthesized parts. They used well-known neural networks as the backbone network such as HRNet \cite{sun2019deep} or XceptionNet \cite{chollet2017xception} with combining their own methods. These existing 2D CNN-based models have good classification performance for single image input. However, these models are not suitable for determining the authenticity of a video because these 2D models are limited to extract only spatial features rather than temporal features.

Even for the same video, fingerprints appearing in a frame vary slightly over time due to external factors such as lighting brightness and camera shake. Contemporary face synthesizing algorithms cannot blend faces by reflecting these fingerprint's subtle changes. Therefore, spatial and temporal interpretation of the extracted edge region features can lead to higher performance in determining the authenticity of a video.

Lately, 3D-CNN models are designed by borrowing the structure of an existing 2D-CNN model for action recognition and medical image analysis \cite{hara2017learning, bui20173d}. 
In particular, a recent face manipulation detection study \cite{qian2020thinking} uses a SlowFast \cite{feichtenhofer2019slowfast}, which is for video feature extraction, as a backbone network. Studies applying these 3D-Models to video-related tasks show remarkable performance in each field. However, all 3D networks of previous studies have structures that receive relatively large input sizes, usually whole frames. On the other hand, our extracted features have a small size, around 20 pixels in height and width. Therefore, it is not appropriate to utilize other existing networks as a backbone for our framework.

Considering these observations, we designed a 3D-CNN model for our framework to classify facial manipulated video. A 3D-LightDenseNet by referring to the DenseNet \cite{huang2017densely, hara2018can} model is devised to obtain richer spatio-temporal information from the processed features. This model takes the edge region features extracted from several consecutive frames within a video as input. The 3D-LightDenseNet has a shallower structure than the existing DenseNet because the size of the extracted edge region is smaller than the entire image. 

Overall, this model behaves in a similar way to the existing DenseNet. This model also consists of Dense Block and Transition Block. We simply convert all 2D-Layers to 3D-Layer. The notable difference is that there is no pooling layer in Dense Block. This is because the size of the initial input is very small. The detailed network structure is shown in Figure \ref{fig:fig8}.


\section{Experiment}

We perform extensive experiments using the DFD dataset \cite{dfd_2019} and 3 face forensic datasets mentioned in Section \ref{aef}. In addition to cross-validation and comparison experiments, in-depth experiments and ablation studies are designed to test the proposed method's robustness and practicality.

\noindent\textbf{Experiment Inputs.} First, we split each dataset into 80\% for training data and 20\% for test data. As with the frame extraction method described in Section \ref{aef}, 24 consecutive frames at random timepoint are extracted per video. Then edge region features are extracted frame-wisely with referring to Section \ref{fe}. The hyper-parameters \(\alpha\) and \(\beta\) in Section \ref{fe} are empirically set to 25, 54, respectively. Extracted features \(F_I\) are shaped 20\(\times\)19\(\times\)24 with 16\(\times\)3 channels, of which 16 refers to the number of facial boundary windows, and 3 indicates RGB. Additionally, we make a difference features \(D_I\) with \(F_I\) and \(B_I\) by referring to the Equ. \eqref{eq6}. Note that \(D_I\) is a post-processed \(B_I\) to reflect the color distribution differences between the background edge region and the face edge region in the classification framework. 

\begin{equation}\label{eq6}
D_{I, i} = F_{I, i} - B_{I}
\end{equation}
\noindent Here, the \(i\) indicates the enumerated edge window. The calculated \(D_I\) is an index indicating how much more focus is placed on the face compared to the background in one image data.

The computed difference features \(D_I\) have the same shape and the same number of channels as the \(F_I\). The final input shape for 3D-LightDenseNet is 20\(\times\)19\(\times\)24 with 96 channels, which 96 indicates sum of the number of channels of the \(F_I\) and those of the \(D_I\).

\begin{table}
\begin{center}
\begin{tabular}{|c| c c c c|}
\hline
\multicolumn{5}{|c|}{Detection Performance (AUC)} \\
\hline\hline
\multirow{2}{*}{Training} & \multicolumn{4}{|c|}{Test} \\
\cline{2-5}
 & FF++ & DFD & DFDC & CelebDF \\
\hline
FF++ & 99.8 & 95.1 & 90.0 & 87.8 \\
DFD & 96.3 & 99.8 & 89.2 & 87.9 \\
DFDC & 98.7 & 93.2 & 98.1 & 88.5 \\
CelebDF & 99.8 & 99.6 & 97.3 & 95.4\\
\hline
\end{tabular}
\end{center}
\caption{Cross-validation performance results.}
\label{table:tb1}
\end{table}

\begin{table*}
\centering
\begin{tabular}{|c|c| c c c c |c|c| c c c c |}
\hline
\multicolumn{12}{|c|}{Detection Performance (AUC)} \\
\hline\hline
\multirow{2}{*}{Method} & \multirow{2}{*}{Training} & \multicolumn{4}{|c|}{Test} & \multirow{2}{*}{Method} & \multirow{2}{*}{Training} & \multicolumn{4}{|c|}{Test} \\
\cline{3-6}\cline{9-12}
& & FF++ & DFD & DFDC & CelebDF & & & FF++ & DFD & DFDC & CelebDF \\
\hline
\multirow{4}{*}{Comp.} & FF++ & 95.4 & 82.3 & 81.6 & 75.7 & \multirow{4}{*}{Resize} & FF++ & 96.5 & 85.7 & 83.3 & 79.9 \\
            & DFD & 83.1 & 95.3 & 82.0 & 79.7 & & DFD & 89.8 & 96.9 & 85.4 & 81.6 \\
            & DFDC & 80.0 & 85.5 & 93.1 & 80.6 & & DFDC & 82.7 & 86.2 & 97.0 & 82.5 \\
            & CelebDF & 84.4 & 83.3 & 82.8 & 86.3 & & CelebDF & 89.4 & 91.1 & 90.2 & 88.9\\
\hline
\end{tabular}
\newline
\caption{Performance results with pixel-level manipulated video. \textit{Comp.} indicates image compression, and \textit{Resize} refers to image downsizing. The results with raw video datasets are in Table \ref{table:tb1}.}
\label{table:tb2}
\end{table*}

\noindent\textbf{Model Details.} The total epoch is set to 1000, and the batch size is set to 20. The learning rate is set as 0.00075 using the Adam \cite{kingma2014adam} optimizer. Additionally, we scheduled the optimizer to multiply 0.5 to the learning rate at every 200 epochs in case of a decaying learning rate. The growth rate is set to 32, and the dense block layers are set to 6, 12, 8 each. The cross-entropy function is used for loss function.

\subsection{Cross-Validation with Various Datasets} \label{CVD}

We firstly design a cross-validation experiment to test our approach in the aspect of robustness to real-world scenarios. In this experiment, we train our model with one dataset and test with another dataset to make scenarios that detect manipulated images with unseen datasets. Table \ref{table:tb1} shows the cross-validation performance result of our approach in terms of AUC(Area Under the receiver operating characteristic Curve).

We figure that cross-validating models trained with Celeb-DF features can obtain more generalized results than models trained with other datasets. This inconsistency may occur because the features extracted from the Celeb-DF dataset have relatively less distinctive differences between unforged images and manipulated images when compared to the other three datasets. We can see that a model trained with a dataset with relatively less prominent features performs well when verifying a dataset with more prominent features. Nevertheless, any experiment result with DFDC and Celeb-DF outperforms other referred face manipulation detection methods to be described later.

\subsection{Comparison with Latest Works}\label{CLW}

\begin{table}
\begin{center}
\begin{tabular}{|c|c c|}
\hline
\multicolumn{3}{|c|}{Detection Performance (AUC)} \\
\hline\hline
Study & FF++ & CelebDF \\
\hline
FWA \cite{li2018exposing} & 93.0 & 64.6 \\
Face X-ray \cite{9157215} & 98.5 & 74.8 \\
Evolution \cite{tolosana2021deepfakes} & 99.5 & 83.6 \\
Two-Branch \cite{masi2020two} & 93.2 & 73.4 \\
\(F^3\) Net \cite{qian2020thinking} & 98.0 & - \\
VST \cite{xu2021visual} & 99.6 & 96.2 \\
SPSL\cite{liu2021spatial} & 96.9 & - \\
BitaNet\cite{ru2021bita} & \textbf{99.8} & \textbf{98.8} \\
\hline
\textbf{Ours} & \textbf{99.8} & 95.4 \\
\hline
\end{tabular}
\end{center}
\caption{Single dataset experiment result compared with contemporary works.}
\label{table:tb3}
\end{table}

\noindent \textbf{Detection Accuracy.} As with the experimental method mainly conducted in other previous studies, we experiment with one specific dataset for both training and testing. The performance results of several existing methods \cite{li2018exposing, 9157215, tolosana2021deepfakes, masi2020two, qian2020thinking, xu2021visual, liu2021spatial, ru2021bita} are compared with those of our proposed method. Experiments are conducted with the FF++ and Celeb-DF datasets, which have been mainly covered in other studies. The summarized performance comparison is in Table \ref{table:tb3}.

The detection results for the Celeb-DF dataset are slightly inferior compared to the latest studies \cite{ru2021bita, xu2021visual}, while results of the FaceForensics++ dataset are superior to other studies. The comparison results of other latest detection frameworks show that the general detection performance itself is also excellent.

\noindent \textbf{Generalization Ability.} We compared our performance results with the latest works \cite{zhangdetecting, li2018exposing,9157215,liu2021spatial, qian2020thinking, xu2021visual} that deal with generalized forgery detection ability for a more precise evaluation of universality. Similar to the cross-validation experiments conducted in other studies, our framework is trained with only the FaceForensics++ dataset to perform comparison experiments. The Celeb-DF dataset and DFDC dataset are used for cross-validation tests. The forged videos in these datasets are made with totally different source data from the those of FaceForensics++ dataset. Table \ref{table:tb4} summarizes the AUC results compared with these existing contemporary works and ours. 

We observe that our framework outperforms the other several approaches in terms of the AUC results. Compared to FF++ results verified with the same dataset, the performance results of cross-validation with DFDC and Celeb-DF are much better than the latest generalization studies.

\begin{table}
\begin{center}
\begin{tabular}{|c| c c c|}
\hline
\multicolumn{4}{|c|}{Detection Performance (AUC)} \\
\hline\hline
\multirow{2}{*}{Study} & \multicolumn{3}{|c|}{Test} \\
\cline{2-4}
 & FF++ & DFDC & CelebDF \\
\hline
3DCNN \cite{zhangdetecting} & 72.22 & 55.02 & 57.32 \\
FWA \cite{li2018exposing} & 93.00 & - & 64.60 \\
Face X-ray \cite{9157215} & 98.52 & 80.92 & 80.58 \\
\(F^3\) Net \cite{qian2020thinking} & 97.97 & - & 65.17 \\
VST \cite{xu2021visual} & 99.60 & 72.45 & 63.47 \\
SPSL\cite{liu2021spatial} & 96.91 & 66.16 & 76.88 \\
\textbf{Ours} & \textbf{99.80} & \textbf{90.03} & \textbf{87.81} \\
\hline
\end{tabular}
\end{center}
\caption{Generalization performance results compared with recent studies. All comparison frameworks are trained using only the FaceForensics++ dataset.}
\label{table:tb4}
\end{table}

\begin{table*}
\centering
\begin{tabular}{|c|c| c c c c |c|c| c c c c |}
\hline
\multicolumn{12}{|c|}{Detection Performance (AUC)} \\
\hline\hline
\multirow{2}{*}{Method} & \multirow{2}{*}{Training} & \multicolumn{4}{|c|}{Test} & \multirow{2}{*}{Method} & \multirow{2}{*}{Training} & \multicolumn{4}{|c|}{Test} \\
\cline{3-6}\cline{9-12}
& & FF++ & DFD & DFDC & CelebDF & & & FF++ & DFD & DFDC & CelebDF \\
\hline
\multirow{4}{*}{w/o BG} & FF++ & 98.5 & 91.2 & 84.8 & 79.3 & \multirow{4}{*}{2D} & FF++ & 95.7 & 92.8 & 86.0 & 82.1 \\
            & DFD & 88.2 & 98.6 & 84.1 & 81.8 & & DFD & 89.5 & 96.2 & 85.4 & 83.3 \\
            & DFDC & 91.1 & 86.7 & 97.3 & 83.0 & & DFDC & 91.7 & 88.5 & 94.6 & 82.8 \\
            & CelebDF & 94.2 & 91.9 & 87.4 & 93.1 & & CelebDF & 95.0 & 94.5 & 91.2 & 89.9\\
\hline
\end{tabular}
\newline
\caption{Performance results of ablation studies. \textit{w/o BG} refers to without background features, and \textit{2D} indicates 2D-variation model of 3D-LightDenseNet. The vanilla results are in Table \ref{table:tb1}.}
\label{table:tb5}
\end{table*}

\subsection{Forgery Detection Against Pixel Manipulation}

\noindent\textbf{Image Compression.} We evaluate whether the proposed detection framework has robust detection capabilities in terms of image compression. The experiment is performed by compressing the image quality of videos in existing datasets. Image compression is performed on all datasets used for training and testing. A low-quality(LQ) compression method referred to by \cite{Rossler_2019_ICCV} is applied to experimental datasets. The test results are in the left part of Table \ref{table:tb2}. 

Compared with Table \ref{table:tb1}, we see that the experimental results that verifying with the same dataset don't get much worse. But, the detection performance that verifying with other datasets are poorer than the original results. Hence, feature extraction at the pixel level is especially affected by image quality in cross-validation experiments.

\noindent\textbf{Image Downsizing.} We additionally test whether the framework shows robust performance even for small-resolution videos. The video used in the experiment is 1/16 (1/4 width, 1/4 height) of the video resolution of the existing dataset. A simple bilinear resizing method is used. The experimental results are shown in the right part of Table \ref{table:tb2}.

The performance deterioration is unavoidable with reduced pixel information like the image compression experiment. However, it shows similar detection performance compared with existing studies referred at Section \ref{CLW}, even with downsized videos tests. Comparing these experimental results with the performance of other existing studies, it appears that our method is relatively more robust to data laundering like the resolution reduction.


\subsection{Ablation Studies}

\noindent\textbf{Effect of Background Features.} We conduct an ablation study to see how much the background edge features affect the detection performance. The experiment evaluates the detection performance by using only the 48 input channels representing the facial edge region features. We change the input channel of the detection model from 96 to 48. Additionally, the numbers of output channels of the convolution layers in the classification model are also proportionally halved. The detection model is then fine-tuned. The test results are presented in the left part of Table \ref{table:tb5}.

The ablation experiment result of the background edge information is worse than the original test result. Especially in cross-validation experiments, AUC results deteriorated by 5-8\%. The experimental results demonstrate that reflecting the features of the background boundary together in the detection framework leads to better generalization performance.

\noindent\textbf{Effect of Time-domain information.} Most of the contemporary existing face manipulation \cite{8682602, li2018exposing, tarasiou2020extracting, liu2021spatial} detection studies do not determine the authenticity of the whole video but in units of frames that exclude the concept of time. We utilize 2D-LightDenseNet to evaluate detection performance with only frames, excluding the concept of time. A 2D-LightDenseNet is a transformed model that replaces all 3D layers of 3D-LightDenseNet with 2D layers. The model takes an input with the dimension of the number of frames removed. The evaluation method for determining the authenticity of an image proceeds with reference to \cite{Rossler_2019_ICCV}. The experimental results are in the right part of Table \ref{table:tb5}.

The performances of the time-domain ablation study are inferior to the experimental results through the original 3D model. It shows a performance drop of about 4-7\% in all cross-validation AUCs.
As a test result shown above, it is important to include the time-domain features when detecting facial tampered videos.

\section{Conclusion}

We tackle the generalized facial forgery detection problem by leveraging the pixel-level difference in the boundary area made by the post-processing procedure of synthesis algorithms. Our work shows the importance of utilizing both face edge information and background edge information for generalized detection abilities. Furthermore, we demonstrate that spatiotemporal interpretation of these edge region features is more advantageous for image authenticity determination. Outstanding performance results from extensive cross-dataset experiments demonstrate that our approach not only performs well with a specific dataset but can also be fully applied in real-world situations.

\vspace{3mm}

\noindent\textbf{Acknowledgements.} This work was partly supported by Institute of Information \& communications Technology Planning \& Evaluation (IITP) grant funded by the Korea government(MSIT) (IITP-2021-2015-0-00742, Grand Information Technology Research Center support program, 50\%) and National IT Industry Promotion Agency of Korea(NIPA) grant funded by the Korea government(MSIT) (No.S0316-21-1006, Healthcare AI Convergence Research \& Development Program, 50\%).

\printbibliography

\end{document}